\documentclass{article} % For LaTeX2e
\usepackage{nips15submit_e,times}
\usepackage{hyperref}
\usepackage{graphicx}

\usepackage{subcaption}

\usepackage{amsmath}
\DeclareMathOperator{\avg}{avg}

\usepackage{url}
\usepackage{caption}

\title{Learning to Communicate in Multi-Agent Reinforcement Learning : A Review}

\author{
Mohamed Salah Zaiem\thanks{Both authors contributed equally}
\\
Ecole polytechnique, Paris\\
\And
Etienne Bennequin\footnotemark[1] \\
Ecole polytechnique, Paris \\
}

\nipsfinalcopy % Uncomment for camera-ready version

\begin{document}

\maketitle
\begin{abstract}
We consider the issue of multiple agents learning to communicate through reinforcement learning within partially observable environments, with a focus on information asymmetry in the second part of our work. We provide a review of the recent algorithms developed to improve the agents' policy by allowing the sharing of information between agents and the learning of communication strategies, with a focus on Deep Recurrent Q-Network-based models. We also describe recent efforts to interpret the languages generated by these agents and study their properties in an attempt to generate human-language-like sentences. We discuss the metrics used to evaluate the generated communication strategies and propose a novel entropy-based evaluation metric. Finally, we address the issue of the cost of communication and introduce the idea of an experimental setup to expose this cost in cooperative-competitive game.
\end{abstract}

\paragraph{Disclaimer} This review has been written in December 2018 as coursework. In a very dynamic field, interesting results may have been found meanwhile, and a few assumptions here may have become outdated. However, we believe that this remains a good start for people discovering this field or trying to get an idea on what has been done about communication in multi-agent reinforcement learning settings. 

\section{Introduction}
In recent years, Multi-Agent Reinforcement Learning has received a lot of interest, with more and more complex algorithms and structures improving the policy of multiple agents in more and more complex environments \cite{comprehensive:survey}. In Multi-agent settings, agents take actions and learn from their rewards simultaneously in the same environment. These interactions could be either competitive \cite{gosilver} or cooperative or a mix of the two. 
This review mainly tackles the case of partially observable multi-agent environments. In real world environments it is rare that the full state of the system can be provided to the agent or even be determined. In this kind of environments, agents may hold very valuable information for the decision making of other agents. In the case of common reward, agents would need to communicate seamlessly with the other agents in order to pass that information, coordinate their behaviours and increase the common reward \cite{Littman:1994}.

Classic independent Q-learning or DQN learning have showed poor performance for this kind of setups \cite{matignon:2012}. Since no communication is involved, other agents learning and acting accordingly are considered part of the environment in these models. The environment is therefore non stationary, and agents fail to converge to an optimal policy. Allowing the agents to share messages and learn what to communicate can significantly improve their adaptability to a given situation.

To allow this information to circulate between agents, there has been a lot of research work on creating communication channels and protocols. These channels can be either continuous, therefore allowing easy derivation and optimization, or discrete with a set of symbols/letters copying human messages' generation. These messages can either be determined in advance by a human operator or be learned by the agents, using the reward as the only learning signal. In this work, we will focus on the latter case.

We will separate two different uses of this framework. They can be roughly explained this way : the first tries to improve agent's ability in a task through communication between agents, while the second one tries to improve the quality of the generated "language" (or communication) and its closeness to natural language through mastering a task.

The first use does not give a primary importance to the format of communication. Information could circulate through handmade communication channels and protocols as explained above. But it also could be latent and implied, through sharing the Deep Q-Networks weights. 

The second use is motivated by the work of linguists on the emergence of language. Word2Vec embeddings arose from applying to huge textual datasets a famous quote of John Rupert Firth
" You shall know a word by the company it keeps." Concerning the rise of language, Wittgenstein \cite{wittgenstein} said that "Language derives its meaning from use". From this quote, a lot of works have been trying to create a language relying on its functionality, and Reinforcement Learning represents a perfect framework for such creation. In other words, language is a way to coordinate and achieve common objectives and cannot be represented by simple statistical associations like Language Models. Moreover, cognitive studies tend to show that feedback is paramount for the language learning process \cite{sachs}.

Emergent communication and language is not a new field. Wagner \emph{et al.} \cite{wagner} have already proposed a thorough review on the progress of the research in this domain. Since the majority of the works presented by them are not Reinforcement Learning related, and since the Deep Q-Networks offered a lot of new possibilities for researchers, we will not present these works in details.    

We will start by defining the useful tools used in the different setups and models described in the review, then we will see the effect of introducing communication channels and algorithms on the performance of agents in multi-agent partially observable settings. We will then inspect the link between the generated languages and natural ones, studying the attempts of researchers to ground the communication sequences into human language. In the final part, we will discuss the metrics used in the experiments proposing a new entropy-based one, and we will propose an new multi-agent environment that could expose a cost in communication between agents. 

\section{Useful tools}
\paragraph{Deep Q-Networks} Mnih \emph{et al.} proposed in 2015 a representation of the Q-function of the policy of an agent using deep neural networks \cite{dqn}.

Assume a single-agent observing at each time step $t$ the current state $s_t$ (fully observable) and taking an action $a_t$ following a policy $\pi$ with the objective of maximizing an expectation of the sum of future $\gamma$-discounted rewards $R_t = r_t + \gamma r_{t+1} + \dots$. In Q-learning, this expectation is represented by the value function of the policy $Q^{\pi}(s, a) = \mathbf{E}(R_t|s_t=s, a_t=a)$, with the optimal value function obeying the Bellman equation :
\begin{equation*}
    Q^*(s,a) = \mathbf{E}\left[ r + \gamma \max_{a'}Q^*(s',a')|s,a \right]
\end{equation*}
In DQN, a neural network is used to represent the value function $Q(s,a,\theta)$ where $\theta$ is the parameters of the network. The training of the network is done using \emph{experience replay}, \emph{i.e.} a dataset of past experiences $e_t = (s_t, a_t, r_t, s_{t+1})$ sampled at random to constitute mini-batches, and the loss function w.r.t. the target $y = r + \gamma \max_{a'}Q(s',a', \theta^-)$ ($\theta^-$ being the parameters of a copy of the network, frozen at the beginning of each update iteration) :
\begin{equation*}
    L(\theta) = \mathbf{E}_{s,a,r,s'} \left[ ( y - Q(s,a,\theta) ) ^2  \right]
\end{equation*}
In the case of multiple agents, this can be extended to \emph{Independent DQN}, in which the Q-network of each agent is updated independently from the others. It has to be noted that this algorithm can lead to convergence problems, since the environment can appear non-stationary to one agent because of the learning of the other agents. Also, this assumes full observability of the state of the environment by all agents, which is a strong assumption in practice.

\paragraph{Deep Recurrent Q-Networks} Hausknecht \& Stone proposed an adaptation of the DQN able to address the issue of a partially observable environment using Recurrent Neural Networks \cite{drqn}.
Instead of approximating a value function $Q(s,a)$, this approximates $Q(o,a)$, where $o$ is the observation made by the agent. Here, $Q$ is computed using a RNN (usually, an LSTM layer), \emph{i.e.} we have at each time step $Q(o_t, h_{t-1}, a, \theta )$. The observations can then be aggregated over time.

\section{Multi-agent Cooperation}
\subsection{Cooperation \& communication}
Most real life situations can be modeled as several or many agents interacting in an environment. These agents can cooperate, compete, or some combination of the two. Tampuu \emph{et al.}'s work presents all of these three possibilities in a game of Pong \cite{pong}.
In their experiment, two agents play the famous Atari game with each other. Each of them can fully observe the state of the environment and tries to find its optimal policy using the independent DQN algorithm. No reward is given during the point. At the end of each point, the player who scored receives a reward $\rho$, and the other one receives a reward $-1$. Then the authors make $\rho$ vary between +1 and -1. When $\rho = 1$, the game is fully competitive, since each player has a full incentive to score. When $\rho=-1$, the game is fully cooperative : the two players are encouraged to keep the ball alive. Their experiment showed that with a competitive rewarding scheme, the number of paddle-bounces between the two players is about a hundred times lower than it is with a cooperative rewarding scheme, which means that in the latter, both players learned how to cooperate in order not to drop the ball.

This experiment shows that the notion of cooperation in a multi-agent environment is close to the notion of shared reward between the agent. If several agents evolving in the same environment share the same common reward, then they have full incentive to cooperate.

However, this does not mean that agents sharing the same reward will have an incentive to communicate. According to most papers in this area, the incentive to communicate actually comes from a partially observable environment. Take for instance the switch riddle : one hundred prisoners arrive in a new prison. The warden tells them that each day, one of them will be chosen at random (among all prisoners, including those having already been picked) and placed in a room containing only a light bulb. This prisoner can observe the state of the light bulb (\emph{on} or \emph{off}) and choose whether to change its state or not. He can also announce that he believes that all prisoners have visited the room. If he is right, all prisoners will be set free, but if he is wrong, they will all be executed. The prisoners can agree in advance on a strategy but as soon as the challenge begin, they will not be able to communicate with each other, neither will they know who is picked to go to the light bulb room.

In this riddle, prisoner-agents cannot observe the full state. They cannot observe anything, unless they are picked to go in the room, in which case they only know that they are the one in the room, along with the state of the switch. However, the common knowledge of all agents covers the full state and allows to solve the riddle. Therefore, the agents have an incentive to communicate, through the light bulb.

The problem of a partially observable multi-agent environment was already addressed in 2004 by Ghavamzadeh \& Mahadevan \cite{hrl}, who used Hierarchical Reinforcement Learning to allow multiple taxi drivers to learn to pick-up as many customers as possible in a finite time.
After executing an action, a taxi driver is able to receive the state of the other cabs. However, communication in this case is not free : there is a arbitrarily predefined cost for getting the information from the other taxi drivers. Thus each agent had to compare the value function without communication with the sum of the value function with communication and the cost of communication. The authors did a comparison between a "selfish multi-agent" baseline, in which the agents do not communicate with each other, and their algorithm for different values of the communication cost. They found that even though the results were always better with their algorithm, the number of satisfied customers decreased when the communication cost increased, tending close to the baseline results. These experiments have to their credit that they managed to quantify the trade-off between partial observation and communication cost, and therefore to quantify the incentive to communicate.

\subsection{Centralised learning and Deep Reccurent Q-Networks}
Some constraints on the nature of the communication can make it non-differentiable, thus invalidating a lot of learning algorithms. A widely used solution to avoid this issue is centralised training for decentralised execution. Agents have a learning period in which they develop their communication schemes along with their policies, which they apply afterwards, at test time, possibly with more constraints on the message.

\paragraph{Learning communication with backpropagation}Sukhbaatar \emph{et al.} \cite{backprop} developed a "Communication Network", that takes as entries all the partial observations of the $M$ agents and outputs the actions. Each layer $i$ of this $N$-layered network has $M$ cells $\lbrace f^i_m \rbrace _{m=1,\dots,M}$. Each cell takes as input $(c_m^i, h_m^i)$ where $h_m^i$ is the output of $f_m^{i-1}$ (with $h_m^0 = g(s_m)$) and $c_m^i = \avg\limits_{m'\neq m}h_{m'}^{i-1}$ is the message coming from other agents (it can also be computed only on a subset of the set of other agents), and outputs :
\begin{equation*}
    h_m^{i+1} = \sigma(C^i c_m^i + H^i h_m^i)
\end{equation*} where $\sigma$ is some non-linearity (as an alternative, $f_m^i$ can also be an LSTM). Finally $f^N_m(c_m^N, h_m^N) = a_m$. The weights of the network are trained all-together using backpropagation from a loss related to the reward on a batch of experiences. Their model proved to heavily outperform a baseline model where communication inside the network occurs in discrete symbol.

After some promising results, the research on centralised learning of communication schemes took a turn with the utilisation of Deep Recurrent Q-Networks. Unlike the independent Deep Q-Network used in Tampuu \emph{et al.}, Deep Recurrent Q-Networks do not assume full observability and do not lead to convergence problems (in independent DQN, one agent's learning makes the environment appear non stationnary to the others).

Foerster \emph{et al.} proposed the two following algorithms deriving from DRQN in which multiple agents learn to communicate in a centralized fashion and sharing the weights of their network. Both make the assumption of a partially observable environment in which agents share the same reward.

\paragraph{Deep Distributed Recurrent Q-Network \cite{DDRQN}} This algorithm derives from a combination of DRQN and independent Q-learning, with three modification : the experience replay is forgotten (as said before, the experience of one agent is made obsolete by the learning of the others), the last action is added as input, and most importantly the weights of the network are shared between all the agents : there is not one Q-network for each agent, but one Q-network in total, allowing more efficient (and way less costly) common learning. Each agent can still behave differently from the others since it gets its own individual inputs. The index of the agent is also given as input, in order to allow each agent to specialise. The Q-function to be learned by the RNN becomes of the form $Q(o_t^m, h^m_{t-1}, m, a_{t-1}^m, a_t^m, \theta)$, with $o_t^m$ and $a_t^m$ respectively the observation and the action of agent $m$ at time step $t$, and $h^m_{t-1}$ the hidden state of agent $m$ after time step $t-1$. The network is trained in the following way : 1) the multiple agents evolve during one episode with an $\epsilon$-greedy policy following the Q-function given by the RNN. 2) The weights of the network are updated following Bellman equations.

\begin{figure}
    \centering
    \begin{subfigure}{.45\textwidth}
    \centering
    \includegraphics[width=.99\linewidth]{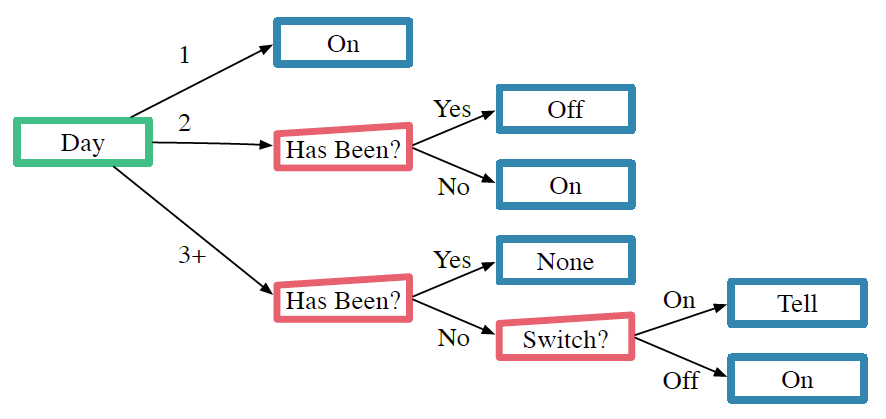}
    
    \caption{\label{fig:switch}Figure~\ref{fig:switch} : Communication protocol developped with DIAL and DDRQN for the resolution of the switch riddle with $n=3$}
    \end{subfigure}%
    \quad \quad \begin{subfigure}{.45\textwidth}
    \centering
    \includegraphics[width=.99\linewidth]{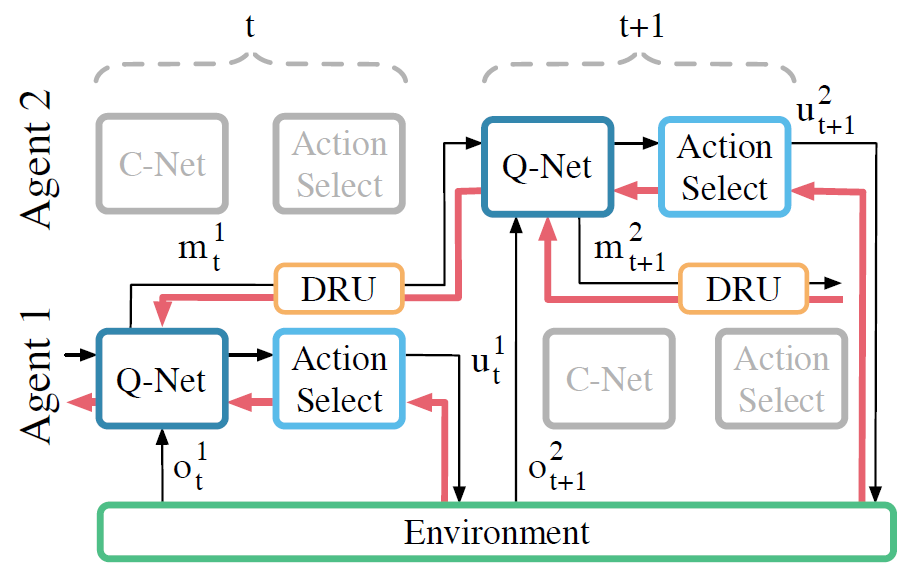}
    
   \caption{\label{fig:DIAL}Figure~\ref{fig:DIAL} : Dial architecture with two agents : red arrows represent backpropagation. Different Q-Net cells are copies of the same shared network.}
    \end{subfigure}

\end{figure}

\paragraph{Differentiable Inter-Agent Learning \cite{DIAL}} At each time-step, in addition to an environment action $a_t^m$, each agent $m$ takes a communication action $c_t^m$. This message is not restricted during the centralised learning but is restricted to a limited-bandwidth channel during the decentralised execution. \\
Here, the Q-network does not only output the Q-function, but also the real-valued (thus differentiable) message $c_t^m$ which is fed at time-step $t+1$ to the Q-networks of the other agents (that still share the same weights). Then two separate gradients back-propagate through the Q-network : 1) the one associated to the reward, coming from the environment. 2) the one coming from the error backpropagated by the recipients of the message in the next turn. Figure~\ref{fig:DIAL} shows the DIAL architecture with two agents. With this model, messages can be updated more quickly to minimize future DQN loss.

Note that in the first model, there is no explicit message fluctuating between the agents. The communication lies in the actions that are taken by the multiple agents and can influence the observations of other agents in the future. But in both algorithms, there is a big deal of implicit communication during the learning, when all agents actually come together and agree on a common strategy, just like the prisoners in the switch riddle.

Both algorithms having been tested on their performance on this switch riddle, we can compare their results. Both algorithms usually converge to the same rewards : $100 \%$ of the Oracle (maximal possible reward) for $n=3$, and $90 \%$ for $n=4$. It is also interesting to see that the same communication protocol is used by both algorithms after convergence for $n=3$ (this protocol is shown in Figure~\ref{fig:switch}). However, DIAL converges about 20 times faster than DDRQN in both cases (after one thousand episodes for DIAL, but 20 thousands for DDRQN). Plus, in the case of $n=4$, even though DDRQN converges in most instances to $90 \%$ of the Oracle, there are a significant number of instances in which we cannot see any convergence after 500 thousands episodes. This is a good argument in favor of differentiable message directly adjusted following downstream DQN loss.

\subsection{Cooperation through public belief}
However, the DIAL method assume the existence of a channel on which agents can communicate freely without affecting the environment. In practice, this is not always the case. Foerster \emph{et al.} \cite{BAD} introduce the notion of \emph{public belief}, which can be defined as the estimation of the probability of possible states (\emph{i.e.} observable or not) knowing all public observations (\emph{i.e.} observations available to all agents) : assuming that at time step $t$, $f_t$ is the discrete set of features of the environment that compose the state $s_t$, and that $f_t^{pub}$ is the subset of features that are commonly known by all agents, we define the public belief $\mathcal{B}_t = P(f_t | f^{pub}_{1,\dots,t})$.

\paragraph{Bayesian Action Decoder \cite{BAD}} The authors proposed the \emph{BAD} algorithm to use the notion of public belief in order to allow agents to learn a cooperative policy. \\
They assume a partially observable environment with no communication channel. At each time step $t$, agent $m$ takes an action $a_t^m$ following a policy $\pi^m$ knowing all its past observations and actions. All agents share a common reward. The training is centralised, which means that each agent knows the policies of all the other agents. \\
The main idea is to generate a virtual third-party agent with \emph{public belief policy} $\pi_{\mathcal{B}_t}$ that conditions on public information and public belief, and therefore can be computed by each agent through a common algorithm. At each time step, this virtual third-party outputs a policy $\hat \pi_t$ that maps all possible private observations made by the agent $m$ acting at this time step to the set of its possible actions. Then, at time step $t+1$, since all agents know the policy $\hat \pi_t$, the action that agent $m$ has taken following this policy gives public information about the private observation he made :
\begin{equation*}
    P(f_t^m | a_t^m, \mathcal{B}_t, f_t^{pub}, \hat \pi_t) \propto \mathbf{1}(\hat \pi_t(f^m_t) = a_t^m)P(f_t^m|\mathcal{B}_t, f^{pub}_t)
\end{equation*}
with $f^m_t$ being the state features observed by agent $m$ at time $t$. This allowing to update the public belief.

The authors chose to apply their algorithm to the cooperative firework-themed playing card game Hanabi. In this game, each player can see the other players' cards but not its own, and can chose to hint a teammate about one of this teammate's cards, or to play one of its own cards, using the hints given by the other players. In the case of a two players game, the score converged after ten thousands games, achieving state of the art results. What is even more interesting is that agents developed their own communication conventions through their actions. For instance, the authors found that an agent hinting "red" or "yellow" is a huge incentive for the other agent to use its newest card at the next action. In this way, the agents learned a crucial ability in the interactions between humans, which is the ability to deduce, from the action of someone, latent information about their situation and observations.

\section{Learning a language}
\subsection{Main models : Referential Games }
As we mentioned earlier, emergent languages and communication is not a new research topic, but the development of Deep Recurrent Q-Networks methods offered a whole new lot of possibilities to language researchers.

A majority of the reinforcement learning setups we will describe in this part rely on referential games. Referential games are a variant of the Lewis signaling game \cite{lewis} and have been often used in linguistics and cognitive sciences. They start by presenting a target object to a the first agent (the speaker). The speaker is therefore allowed to send a message describing the object to a second agent (the listener). Finally, the listener tries to guess the target shown to the speaker among a list of possible candidates. The communication is successful, and thus the reward positive, if the listener picks the correct candidate. 

Let us describe more precisely a specific model, the one developed in \cite{lazaridou:17}. Target objects and candidates are sampled pictures from ImageNet. They represent basic concepts like cats, cars... classified among 20 general categories (animal, vehicle...). As it receives raw pixel input, the speaker agent has to detect by itself the features it could send to the listener. Other papers like \cite{lazaridou:18} tried to use disentangled inputs with attribute-based object vectors. 

Two images $i_R $ and $i_L$ are drawn randomly from the set. One of them is the target $t$. The sender knows which one is the target, and generates a message following the policy $s(\theta_s (i_L, i_R, t )) $ from a vocabulary V of size K. The listener tries to decode the message and guess the target image following a policy $r(i_L, i_R, s(\theta_s (i_L, i_R, t ))) \in \{L,R\} $. If $r$, the guess of the listener, is equal to $t$, they both receive a positive reward, otherwise, they receive 0 as a reward.

\subsubsection{Vocabulary}
The size of the vocabulary has a huge influence on the performance of communicating agents. Foerster \emph{et al.} \cite{DIAL} started with sending simple 1-bit message, with their environments needing "small" information (for instance, the switch riddle or finding the parity of a number), but the more complex the problems get, the more vocabulary is needed to ensure informative messages. Lazaridou \emph{et al.} \cite{lazaridou:17} tried single-symboled messages with a vocabulary size going from 0 to 100. Later research, inspired by natural language, suggested the use of LSTM based encoder-decoder architectures to generate sequences \cite{havrylov} \cite{lazaridou:18}. An important point is that these symbols have no \emph{a priori} meaning. They do get linked to visual or physical attributes only during learning. 

One of the main benefits of generating sequences is allowing the speakers to generate languages ensuring compositionnality. With a compositional language, the speaker would be able to describe a red box, if he had prior suffixes or prefixes for "red" and "box" given the training set, even if no "red box" is seen during training. 

\subsubsection{Q-Models}
Each agent (speaker and listener) has an input network that takes the incoming inputs (images, messages, previous actions...) then embeds them using Recurrent Neural Networks or simple shallow Feedforward networks. Lazaridou \cite{lazaridou:17} observed that producing these features with Convolutional Neural Networks in the case of images input yielded better features and therefore better results in the guessing phase.

To generate a sequence of symbols, Lazaridou \emph{et al.} \cite{lazaridou:18} chose a recurrent policy for the decoder \cite{drqn}. The purpose being that the RNN will be able to retain information from past states and use them to  perform better on long sequences generation. Evtimova \emph{et al.} \cite{evtimova} added an attention mechanism during the generation of messages, reporting that it improved the communication success with unseen objects, as the embedding layers focus on recognizable objects. 

To make it clearer, from the images $i_L$ and $i_R$, a convolutional network outputs a hidden state $h_{LR}$ that is input to a single-layer LSTM decoder (the speaker's recurrent policy). This layer generates a message $m = g(h_{LR}, \theta_g ) $. The listener applies the single-layer LSTM encoding for the message (since it is also dealing with sequence input) producing an encoding $z = h(m, \theta_h)$. Coupled with encoding for the candidate images, it predicts a target $t$ following similarity measures between candidates $u \in U$ and encoded vector $v$. 

All the weights of both the speaker and listener agents are jointly optimized using the choice of the agents as the only learning signal. Since their task is different, no weights are shared between the trained agents. 

\subsubsection{Variants}
We described above a popular model in language emergence, but since the emergent languages literature has become consistent, there is a few interesting variants which we will discuss here. \\
A first interesting one is proposed in the work of \cite{jorge:16} and \cite{cao}. To get closer to human-like interactions, they implemented multi-step communication between the agents. In \cite{jorge:16},  the task is very similar, recognizing target images (celebrities in this case), but both agents generate messages. The guessing agent starts the communication given the candidates by asking "questions". The answering agent, who knows the target, answers these questions by yes or no, like in the "Guess Who" game. The best results were obtained using two rounds of question-answer, and their analysis shows that the questioning agent targets different features in Question 1 and Question 2, getting more information. 

Cao \emph{et al.} \cite{cao} also tested multi-step communication but with important changes. It is no longer about referential games, but a negotiation to trade items having different utilities for the two parties. Two channels are implemented, the first one is made to communicate information (like the channels seen earlier), the second one is a concrete trade offer that can be either accepted or rejected by the other agent. When the agents cooperate, \emph{i.e.} the reward is common, the messages indicate to the other agent which items are most valuable and their value. The results show that the "linguistic" channel helps reach a Nash equilibrium and reduces the variance in joint optimality, making the trade system more robust.

\subsection{Results}
The first question this part should answer is whether the communication is successful, \emph{i.e.} whether the agents complete their task. In almost all the papers we have read, agents learn to coordinate almost perfectly and reach very high accuracies. 

For instance, in the work of Lazaridou \emph{et al.} \cite{lazaridou:17}, depending on the model, agents reach a $99\%$ success rate, while the best ones do not fail at choosing the right target. The main differences concern the number of episodes needed before convergence. Best models are the ones that encode best the input images, thus facilitating message generation by a better representation/classification of the described objects.

In \cite{lazaridou:18}, the higher number of distractors (\emph{i.e.} non target inputs to the listener) slightly reduces the accuracy ($98.5\%$). We have to note that success rate increases highly with the size of the alphabet at the beginning, before stabilizing when the vocabulary is large enough to answer the task properly. Table~\ref{tab:lazaridou-results} shows that influence. Lexicon size denotes the effective number of unique messages used by the speaker agent. 

\begin{table}
\center
\begin{tabular}{|c|c|c|r|}
  \hline
 Max length & Alphabet Size & Lexicon size & Training accuracy  \\
  \hline
  2 &10 &31 & 92.0\%   \\
  \hline
  5  & 17 & 293 & 98.2\%  \\
  \hline 
  10 & 40 & 355 & 98.5\% \\
  \hline
  
\end{tabular}
\caption{\label{tab:lazaridou-results}Results of Lazaridou \emph{et al.} 2018 \cite{lazaridou:18}}
\end{table}

\subsection{Link with Natural Language}
As we said in the introduction, research work in multi-agent communication and language emergence did not stick to improving the behaviour of agents facing partially-observable environments. Apart from getting closer to optimal policies, emergent communication has been seen as a way to tackle Natural Language Processing issues. Roughly put, if machines are able to generate a language conveying the same basic functions as human language, an eventual translator from "machine" to human could solve a lot of unsolved NLP tasks. 

Therefore, two challenges arise. The first is interpreting the generated messages. The second is one grounding the emergent language into natural language. Some works have been trying to link the generated messages to human communication, going until twisting the tasks to force that closeness.

\subsubsection{Interpretability}
Starting from rough and noisy messages, interpretation is a rough task. In a lot of cases, researchers fail to understand completely which information the agents are passing and stick to formulating vague hypotheses about meanings. 
 Figure~\ref{fig:conversations} shows the attemps of Jorge \emph{et al.} \cite{jorge:16} to understand the meaning of Question A and B (questions asked by the questioning agent). Looking at the table, it seems that A checks whether the top of the photo is coloured or not(dark or white hair). The operation is even harder with sequences, Table 3 shows the top messages for each object(an object is identified by its shape and color) in the work of Choi \emph{et al.} (ICLR). 

\begin{figure}
\centering
\includegraphics[width=0.5\textwidth]{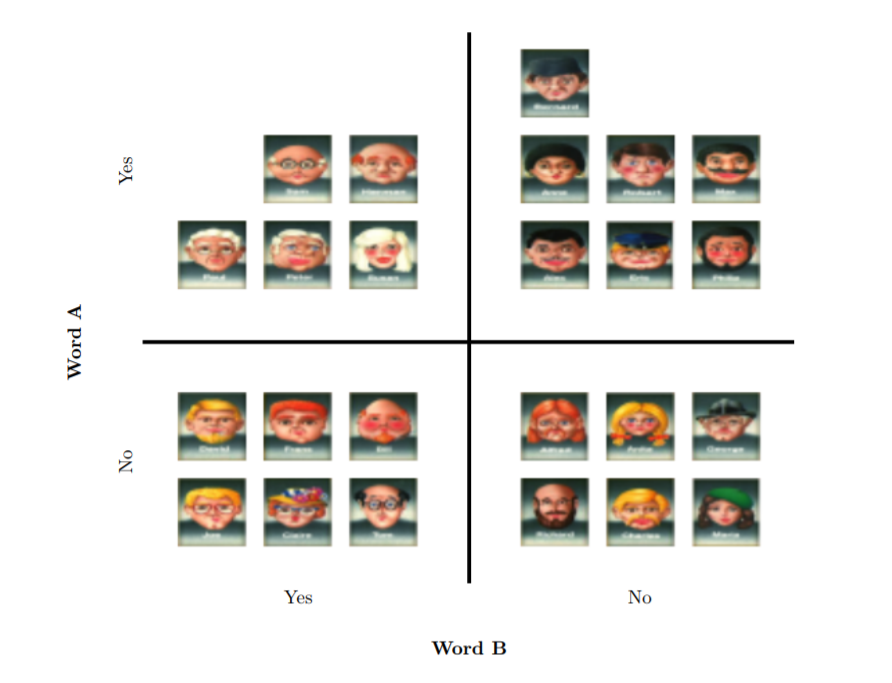}
\caption{\label{fig:conversations}Figure~\ref{fig:conversations} : Samples presented according to the answer they triggered to two questions \cite{jorge:16}}
\end{figure}

\begin{figure}
\centering
\includegraphics[width=1\textwidth]{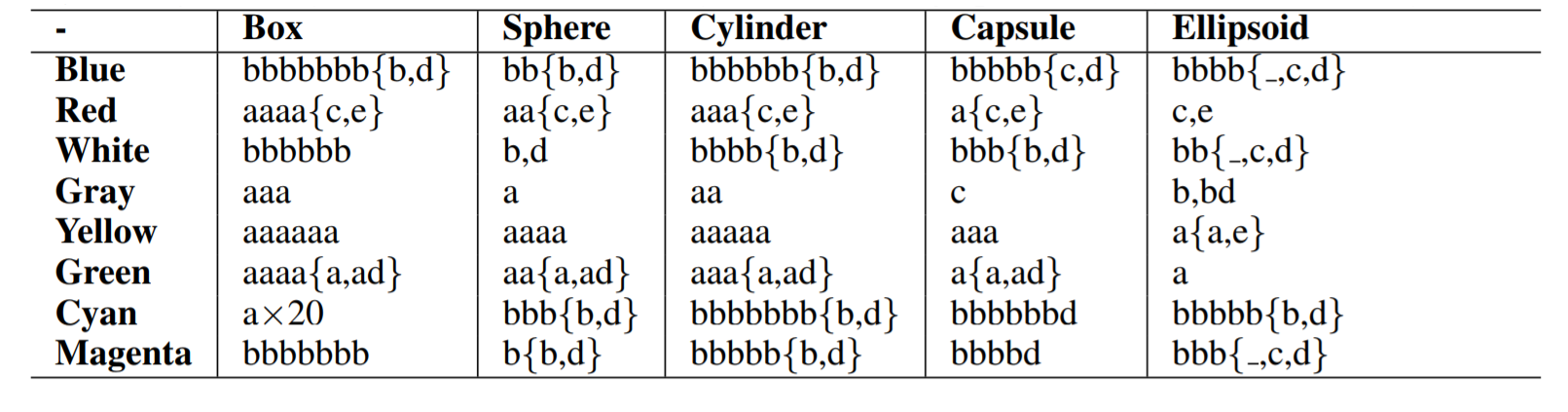}
\caption{\label{fig:conversations}Table 3 : Most common sequences describing objects. (The letters between brackets represent common alternatives for the last letter) \cite{choi}}
\end{figure}

Lazaridou \emph{et al.} \cite{lazaridou:17} tried to estimate closeness to human representations. They used the purity index to assess the quality of "message clusters", \emph{i.e.} the images represented by the same message by the speaker agent. We compare these clusters with the initial ones (the groups englobing the concepts depicted in the images, Dog $\in$ Animals for instance). The purity of a clustering is the proportion of category labels in the clusters that agree with the respective cluster majority category. In other terms, for a given message, we take the label having a majority of elements represented by that message, and check how many images do not pertain to that label but are still represented by the same message. With CNN encoding, and 100 symbols in the vocabulary (43 of them are effectively used) their model reaches $41\%$ purity. 

Choi \emph{et al.} \cite{choi} evaluate closeness to human language by assessing the ability of agents to compose meanings. In this purpose, they use zero-shot evaluation. In their setup, the listener has to guess an image among a set of 4 candidates. Each image is generated randomly given its shape (cylinder, cube...) and color. Zero-shot evaluation consists in showing the trained speaker an unseen combination of shape/color, with previously (multiple times) seen color and shape separately. The point is to evaluate his ability to convey the message of an unseen element composing the knowledge he had on its elements. Communication accuracies do not vary much between seen and unseen objects and stay very close to 1, around 0.97. The emerged language seems able to compose since it succeeds in describing unseen objects.

\subsubsection{Grounding it into Natural Language}
In order to get easy interpretable languages, researchers have tried to compel the agents to generate messages closer to human representations.In \cite{lazaridou:17}, they introduced a few changes to the referential game. Instead of presenting the same image to the speaker and the listener (among other candidates for the latter), they presented an image pertaining to the same concept. For instance, during training, they showed a dog to the speaker, and then the listener had to pick the picture of another dog among a list of non-dog candidates. The point of this manipulation is to force the speaker to send the message "dog", like a human would do, and not send messages on the color of the background, or the luminosity, for instance. This change induced a slight increase in purity reaching $45\%$ . Furthermore, to force even more the speaker to classify "humanly" the images, they alternate its learning between playing the game (generating messages to the listener) and a classic image classification task. 

\section{Discussion and propositions}
\subsection{Lack of evaluation metrics}
The accuracy of the choice is a natural and relevant metric to evaluate the communication's quality, and thus the quality of the language generated. But the metrics used to assess the link with natural language are completely biased by the human macro-representation of raw images. To be clearer, the purity index introduced above is not a valid evaluation of interpretability. We believe that forcing the agent to describe macro-representations(dog, car..) instead of raw image descriptions ( color, background, luminosity..), especially given the lack of prior information on these macro-representations, is not a natural way to solve the task for the agents, and it may harm their communication success. 

However, we believe that it would be interesting to develop an evaluation metric for a learned communication protocol that would be independent from human assumptions as well as from its effectiveness in the execution of a singular task. Zero-shot evaluation for compositionnality is indeed relevant in this perspective, but it cannot capture all the features that make a language relevant.

In 2003, Prokopenko \& Wang studied the entropy on the public belief in a setting with multi-agent coordination \cite{entropy}. We can put that in relation with the results of the Bayesian Action Decoder on the Hanabi game, which shows that the entropy of the public belief (\emph{i.e.} the posterior on all possible states) decreases as the algorithm develops useful communication conventions. In the same spirit, we introduce here the idea of an evaluation metric on a language or communication protocol, independently from the task to be completed by the agents.

\paragraph{Language-Entropy Evolution} Let us assume a two-agent environment, Agent \emph{Smith} being the speaker and Agent \emph{Lee} being the listener. At each time step $t$, \emph{Smith} observes the full state $s_t \in \mathcal{S}$ and sends to \emph{Lee} a message $m_t \in \mathcal{M} $. \emph{Lee} then computes the probability $p_t(s)$ of each possible state $s$ given all messages $m_{1,\dots,t}$. We evaluate the entropy of this probability distribution :
\begin{equation*}
    H(m_1, \dots, m_t) = - \sum\limits_{s\in \mathcal{S}}p_t(s) \log(p_t(s))
\end{equation*}
The comparison between $H(\dots, m_{t-1})$ and $H(\dots, m_t)$ gives a measure of the uncertainty that has been lifted by message $m_t$, in other words how discriminant message $m_t$ is in this context.

Note that the probabilities computed by \emph{Lee} can be on the action set instead of the state set. Then the entropy measures the uncertainty about which action to take instead of the uncertainty about which state the agents are in. In some applications where it is uneasy to compute a probability distribution, this alternative might be more relevant.

We believe that this metric can be useful in the evaluation of learned languages and communication protocol.

EDIT : 
This idea has been explored in the work of Lowe \emph{et al.} \cite{lowe} , along with other interesting evaluation metrics. 

\subsection{Cost of communication}
\begin{figure}
\centering
\includegraphics[width=0.35\textwidth]{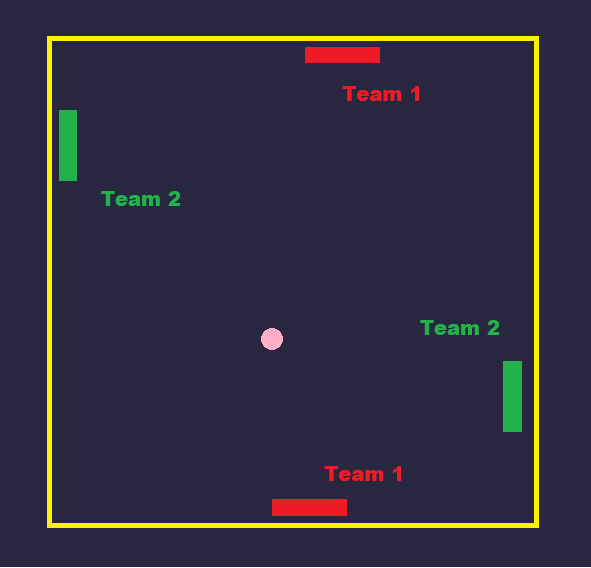}
\caption{\label{fig:conversations}Figure~\ref{fig:conversations} : 4 Players 2 vs 2 Pong Game}
\end{figure}
We also noticed that an interesting study could take place on the possible cost of communication in these systems. Apart from the obvious computational cost, we were thinking about environments with competitive groups of cooperative agents. 

In this context, we have been thinking about a game setting where we would observe a cost on communication in a cooperation/competition setup. The game is represented in Figure~\ref{fig:conversations}, the goal would be to make teams of two agents play against each other a 4-players Pong setting. Multiple cases could be tested : 
\begin{itemize}
    \item No communication channel.
    \item The communication between members of the same team is not heard by others. 
    \item Communication is public, or communication is public for one team and secret for the other.
\end{itemize}
We believe that these experiments could lead to a better understanding of the cost of communication in an environment where sending a message has an indirect influence on the state of the environment, through the reaction of the opposite team.

\section{Conclusion}
We reviewed the main recent algorithms addressing the issue of communication and cooperation between multiple agents in a Reinforcement Learning environment. We also presented the current efforts in interpreting the language learned by these algorithms, and proposed a novel metric to evaluate the relevance of a generated communication strategy.

In the future, we want to implement the 4-players pong environment with the multiple cases described above in a personal project to see how the policies of the teams evolve depending on the constraints on their communication protocol.

\bibliographystyle{unsrt}
\bibliography{mybib}

\end{document}